\documentclass[10pt,twocolumn,letterpaper]{article}
\usepackage{iccv}
\usepackage{times}
\usepackage{epsfig}
\usepackage{graphicx}
\usepackage{amsmath}
\usepackage{amssymb}
\usepackage{amstext,lipsum,mathtools}
\usepackage{afterpage}
\usepackage{caption}
\usepackage{subcaption}
\usepackage{multirow}
\usepackage{xcolor}
\usepackage{eufrak}
\def\G{$\mathfrak{G}$ }
\def\R{$\mathfrak{R}$ }

\usepackage[breaklinks=true,bookmarks=false]{hyperref}

\iccvfinalcopy 

\def\iccvPaperID{8992} 

\begin{document}

\title{WarpedGANSpace: Finding non-linear RBF paths in GAN latent space}

\author{Christos Tzelepis, Georgios Tzimiropoulos, Ioannis Patras\\
Queen Mary University of London\\
Mile End road, E1 4NS London, UK\\
{\tt\small \{c.tzelepis, g.tzimiropoulos, i.patras\}@qmul.ac.uk}
}

\maketitle

\begin{abstract}
This work addresses the problem of discovering, in an unsupervised manner, interpretable paths in the latent space of pretrained GANs, so as to provide an intuitive and easy way of controlling the underlying generative factors. In doing so, it addresses some of the limitations of the state-of-the-art works, namely, a) that they discover directions that are independent of the latent code, i.e., paths that are linear, and b) that their evaluation relies either on visual inspection or on laborious human labeling. More specifically, we propose to learn non-linear warpings on the latent space, each one parametrized by a set of RBF-based latent space warping functions, and where each warping gives rise to a family of non-linear paths via the gradient of the function. Building on the work of~\cite{voynov2020unsupervised}, that discovers linear paths, we optimize the trainable parameters of the set of RBFs, so as that images that are generated by codes along different paths, are easily distinguishable by a discriminator network. This leads to easily distinguishable image transformations, such as pose and facial expressions in facial images. We show that linear paths can be derived as a special case of our method, and show experimentally that non-linear paths in the latent space lead to steeper, more disentangled and interpretable changes in the image space than in state-of-the art methods, both qualitatively and quantitatively. We make the code and the pretrained models publicly available at: \url{https://github.com/chi0tzp/WarpedGANSpace}.
\end{abstract}


\section{Introduction}\label{sec:intro}
    
    \begin{figure}[t]
        \centering
        \begin{subfigure}[b]{0.49\textwidth}
            \centering
            \includegraphics[width=\textwidth]{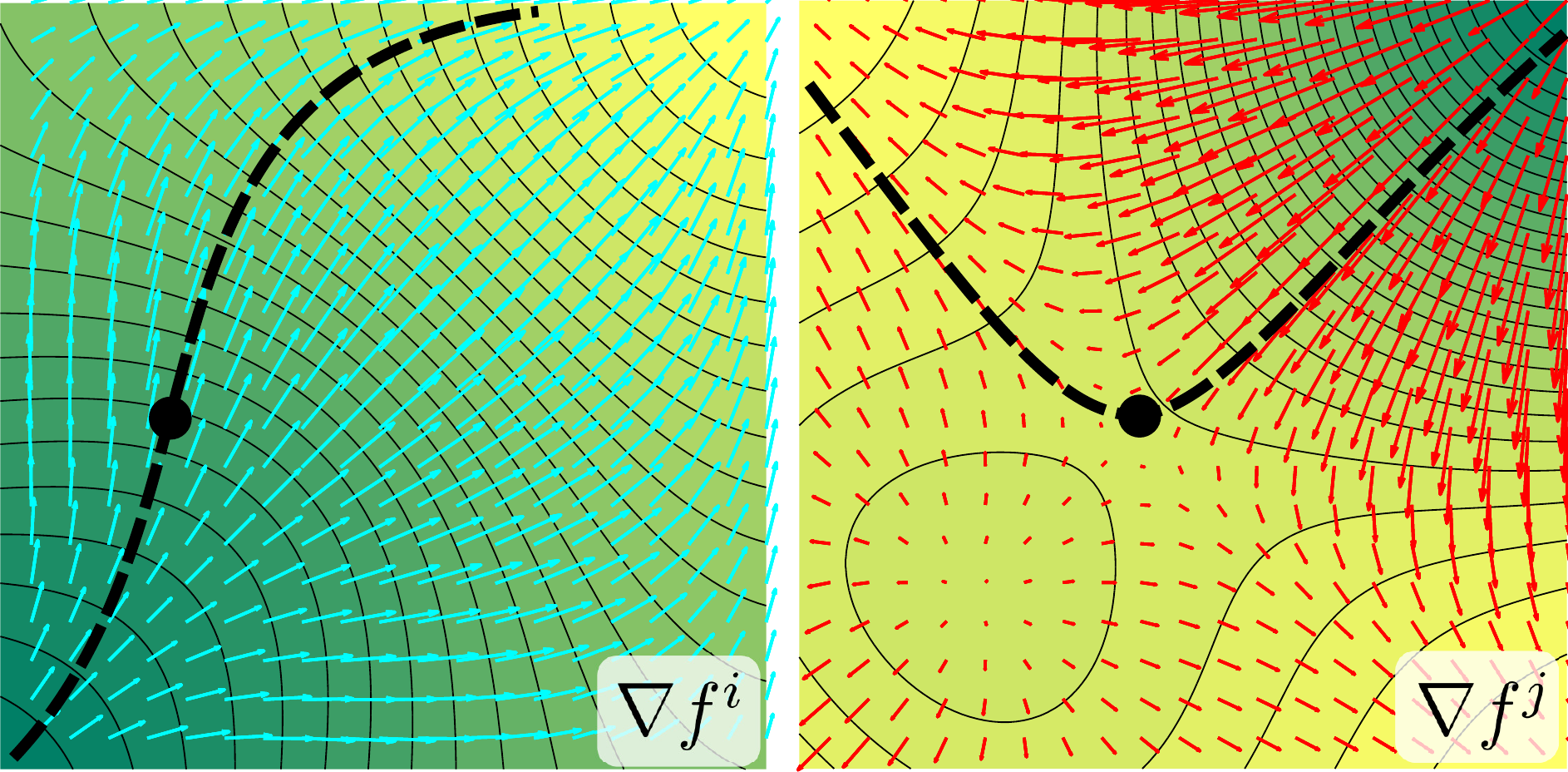}
            \caption{}
        \label{subfig:space_warping}
        \end{subfigure}
        \vfill
        \begin{subfigure}[b]{0.49\textwidth}
            \centering
            \includegraphics[width=\textwidth]{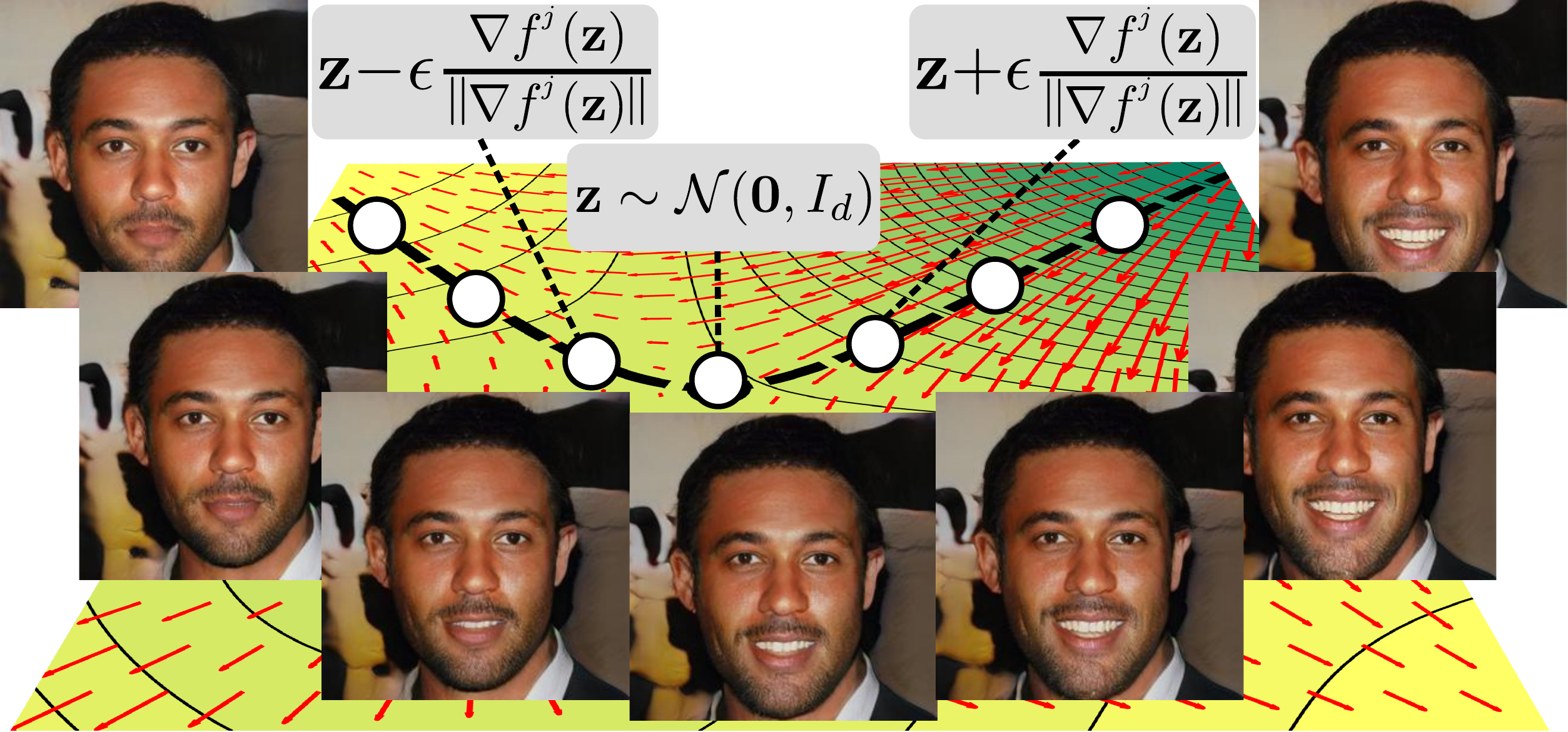}
            \caption{}
            \label{fig:demo}
        \end{subfigure}
        \caption{(a) Warpings of vector space $\mathbb{R}^d$ due to two RBF functions, $f^i$ and $f^j$, lead to different non-linear paths in $\mathbb{R}^d$ for any given $\mathbf{z}\in\mathbb{R}^d$ (dashed bold lines) via their gradients, $\nabla f^i$ and $\nabla f^j$. Solid black lines represent isohypses of the warpings and the colored vectors represent the vector fields induced by their gradients. (b) Illustration of a non-linear path due to warping $f^j$, starting from a latent code $\mathbf{z}$ and moving along the gradient $\nabla f^j$ by steps of magnitude $\epsilon$.}
        \label{fig:fig_1}
    \end{figure}
    
    Generative Adversarial Networks (GANs)~\cite{goodfellow2014generative} have emerged as the leading generative learning paradigm, exhibiting clear superiority in terms of the quality of generated realistic and aesthetically pleasing images~\cite{sngan_miyato18iclr,biggan_brock19iclr,proggan_karras18iclr,stylegan1_karras19cvpr,stylegan2_karras20cvpr}. However, despite their generative efficiency, GANs do not provide an inherent way of comprehending or controlling the underlying generative factors. To address this, the research community has directed its efforts towards studying the structure of GAN's latent space~\cite{unsupervised_radford16iclr,shen2020interpreting,stylegan1_karras19cvpr,voynov19rpgan,denton2019detecting,yang19semantic,bau19iclr,voynov2020unsupervised,xiao2018elegant,goetschalckx2019ganalyze,jahanian20iclr,spingarn2021gan,plumerault20iclr,ganspace_erik2020neurips}. These works study the structure of GAN's latent space and attempt to find interpretable directions on it; that is, directions sampling across which are expected to generate images where only a few (ideally one) factors of variations are ``activated''. Meaningful human-interpretable directions can refer to either domain-specific factors (e.g., facial expressions~\cite{unsupervised_radford16iclr}) or domain-agnostic factors (e.g., zoom scale~\cite{jahanian20iclr,plumerault20iclr,spingarn2021gan}). 
    
    Several methods adopt a supervised learning framework, and discover directions in the latent space that align well to factors controlled by supervision. In this line of research, \cite{interpreting_shen20cvpr,goetschalckx2019ganalyze,stylegan1_karras19cvpr} supervision is in the form of labels assigned to the generated images, either by explicit human annotation, or by the use of pretrained semantic classifiers. Recent works, such as~\cite{spingarn2021gan,jahanian20iclr,plumerault20iclr}, steer the directions in the latent space so as to align well with controllable manipulations in the image space (e.g., zoom). Those works are limited by the fact that the factors are assumed to be known and by practical issues in generating the supervisory signals.
    
    Another line of research imposes unsupervised constraints in the directions in the latent space. GANSpace~\cite{ganspace_erik2020neurips} performs PCA on deep features at the early layers of the generator and finds directions in the latent space that best map to those deep PCA vectors, arriving at a set of non-orthogonal directions in the latent space. Similarly to other methods, this is a very demanding training process that requires drawing large numbers of random latent codes and regressing the latent directions. Similarly, Voynov and Babenko~\cite{voynov2020unsupervised} proposed an unsupervised method to discover linear interpretable latent space directions. While the unsupervised learning framework has interest, current works make the hard assumption that the discovered directions are isotropic in the latent space, leading to linear paths. Furthermore, despite the fact that these works lead to more complex directions, compared to methods that do not use any optimization at all (e.g.,~\cite{jahanian20iclr,spingarn2021gan}), the evaluation of the obtained results are either left to subjective visual inspection (e.g.,~\cite{ganspace_erik2020neurips}) or relies on laborious human labeling~\cite{voynov2020unsupervised}.
    
    \begin{figure*}[t]
        \centering
        \includegraphics[width=\textwidth]{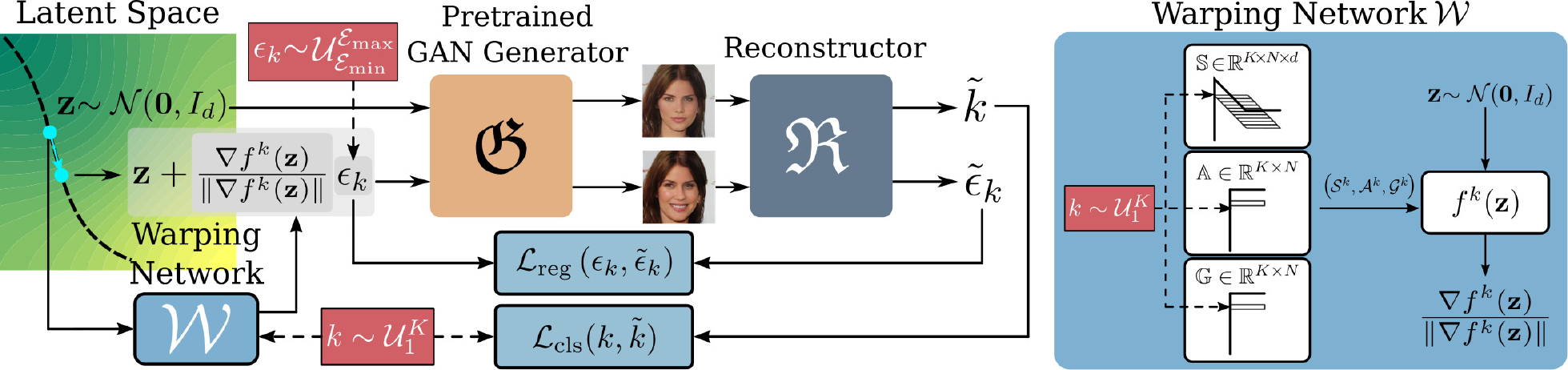}
        \caption{Overview the proposed method: A latent code $\mathbf{z}\sim\mathcal{N}\left(\mathbf{0},I_d\right)$ is shifted by a vector induced by a warping function $f^k$ implemented by the warping network $\mathcal{W}$ after choosing the corresponding support set $\mathcal{S}^k$, weights $\mathcal{A}^k$, and parameters $\mathcal{G}^k$. The pair of latent codes, $\mathbf{z}$ and $\mathbf{z}+\epsilon_k\frac{\nabla f^k(\mathbf{z})}{\lVert\nabla f^k(\mathbf{z})\rVert}$, are then fed into the generator \G in order to produce two images. The reconstructor \R is optimized to reproduce the signed shift magnitude $\epsilon_k$ and predict the index $k$ of the support set used.}
        \label{fig:overview}
    \end{figure*}
    
    In this work, we propose to learn non-linear warping functions on the latent space, each one parametrized by a set of RBF-based latent space warping operations, and where each warping function $f^k$ gives rise to a family of non-linear paths via its gradient. More precisely, at each latent code $\mathbf{z}\in\mathbb{R}^d$, the gradient of the warping function $\nabla f^k(\mathbf{z})$ gives the direction along the $k$-th family of paths -- clearly, the gradient of $f^k$ is not isotropic in $\mathbb{R}^d$, giving rise to non-linear paths. An example is shown in Fig.~\ref{fig:fig_1}, where two RBF-warping functions, $f^i$ and $f^j$, are depicted together with two distinct non-linear paths. Building on the work of~\cite{voynov2020unsupervised}, that discovers linear paths, we optimize the trainable parameters of the RBFs, so as that images that are generated by codes along paths of different families, $f^k$, are easily distinguishable by a discriminator network (Fig.~\ref{fig:overview}) -- this leads to easily distinguishable image transformations, such as pose and facial expressions in facial images (Fig.~\ref{fig:demo}). We show that~\cite{voynov2020unsupervised}, which learns linear paths, can be derived as a special case of our method and we perform extensive comparisons with state-of-the art methods both qualitatively and quantitatively. 
    
    For a quantitative evaluation, we propose to utilize trained classifiers that assign attributes to the generated images and propose a framework that monitors the correlation between paths in the latent space, and the corresponding changes/paths in the attribute space so as to determine how correlated are paths along certain warping functions to certain attributes. We experimentally show that the proposed non-linear paths in the latent space lead to more disentangled and more interpretable changes in the image space than in state-of-the art methods. In addition, we show that for paths of the same length in the latent space, our method is able to produce much larger changes in the attribute space in comparison to the linear one, i.e., the generated attribute paths are much more steep, and that we are able to generate larger attribute changes before the quality of the generated images deteriorates. 
    
    The main contributions of this paper can be summarized as follows:
    \begin{itemize}
        \item We propose an unsupervised and model-agnostic method for discovering non-linear interpretable paths on the latent space of pretrained GANs by using RBF-based warping functions. We derive the case of linear paths as a special case and learn a set of such warping functions so that the corresponding image transformations are distinguishable to each other.
        \item We propose a quantitative evaluation protocol for measuring the interpretability/disentanglement of paths in the latent space, by analysing the corresponding changes to attributes in the generated images, as those are measured by pretrained semantic classifiers (e.g., pretrained face attribute networks).
        \item We apply our method to four pretrained GANs (i.e., SN-GAN~\cite{sngan_miyato18iclr}, BigGAN~\cite{biggan_brock19iclr}, ProgGAN~\cite{proggan_karras18iclr}, and StyleGAN2~\cite{stylegan2_karras20cvpr}) and compare our non-linear paths to linear ones~\cite{voynov2020unsupervised,ganspace_erik2020neurips}, both qualitatively and quantitatively. We show that in comparison to state-of-the-art, our method produces steeper, more disentangled, and longer paths in the attribute space.
    \end{itemize}

\section{Related Work}\label{sec:related_work}
    
    \paragraph{Disentanglement in generative learning} A disentangled representation in the context of generative learning can be defined as one where single latent units are sensitive to changes in single generative factors, while being relatively invariant to changes in other factors~\cite{bengio2013representation}. Imposing disentanglement in the latent space of a generative learning method has drawn significant attention by the research community in recent years. These works typically refer to the notion of a disentangled latent space~\cite{infogan_chen2016neurips,betavae_higgins17iclr,oogan_liu20aaai,hfsynthesis_lee20,ramesh2018spectral,xing2019unsupervised}, in the context of either VAE (e.g.,~\cite{xing2019unsupervised,betavae_higgins17iclr}) or GAN (e.g.,~\cite{infogan_chen2016neurips,oogan_liu20aaai}) and they typically try to improve the architectures and the training protocols of standard generative methods in order to obtain latent spaces where generative factors are disentangled. While these works provide comprehensive theoretical insights, they are typically applied to toy or low-resolution datasets and exhibit inferior results in terms of generation quality and diversity compared to state-of-the-art GANs, such as ProgGAN~\cite{proggan_karras18iclr} or StyleGAN2~\cite{stylegan2_karras20cvpr}. 
    
    \paragraph{Discovering interpretable paths in pretrained GAN generators} Since the early days of GANs, it has been shown that the GAN latent space often exhibits semantically meaningful vector space arithmetic. Radford et al.~\cite{radford2015unsupervised} showed that there exist latent directions corresponding to adding smiles or glasses on faces. This paved the way for the development of methods that would facilitate image editing and has since received significant research attention. Some works~\cite{goetschalckx2019ganalyze,shen2020interpreting,stylegan1_karras19cvpr} require explicit human-provided supervision to identify interpretable directions in the latent space. More specifically,~\cite{shen2020interpreting,stylegan1_karras19cvpr} use classifiers, pretrained on the CelebA dataset~\cite{celeba_liu15iccv}, in order to predict certain face attributes. These classifiers are then used to produce pseudo-labels for a large number of generated images and their latent codes. Based on these pseudo-labels, a separating hyperplane is learned in the latent space giving rise to a direction that captures the corresponding attribute. Plumerault et al.~\cite{plumerault20iclr} also solve an optimization problem in the latent space for maximizing the score of the pretrained model to predict image memorability and then find the directions that increases memorability. By contract to the above works, our method is trained in an unsupervised manner.
    
    Some recent works~\cite{jahanian20iclr,plumerault20iclr,spingarn2021gan} seek those vectors in the latent space that correspond to controlled image augmentations such as zoom or translation. While these approaches have interest, they can find only the directions capturing the transformations that they have been trained on. By contrast, our method can discover non-linear paths that correspond to more complex generative factors (e.g., skin color, age, etc.).
    
    Finally, our method is closely related to those of~\cite{voynov2020unsupervised,ganspace_erik2020neurips}, since we are also learning a set of interpretable paths in an unsupervised and model-agnostic manner. More specifically, Voynov and Babenko~\cite{voynov2020unsupervised} optimize a set of linear interpretable directions, modeled by a set of vectors in the latent space, and they evaluate the performance of their method using the judgements of eleven human assessors. GANSpace~\cite{ganspace_erik2020neurips} is trained in an unsupervised manner in order to discover meaningful directions by using PCA on deep features of the generator. This method seeks linear directions in the latent space that best map to those deep PCA vectors, and results in a set of non-orthogonal directions. Similarly to other methods discussed above, it also requires a very demanding training procedure (drawing random latent codes and regressing the latent directions), while they provide only qualitative evaluation results. 
        
    In contrast to these works, our method discovers non-linear paths in the latent space of a pretrained GAN generator in an unsupervised manner. Moreover, in order to lift the obvious limitations introduced by manual labeling of the discovered paths, we propose a quantitative and automatic evaluation protocol that obtains the most interpretable paths in terms of correlation with a certain number of attributes.

\section{Proposed Method}\label{sec:proposed_method}
    
    In this section, we present our method for discovering $K$ non-linear interpretable paths on the latent space of a pretrained GAN generator, by learning $K$ warping functions, $f^1,\ldots,f^K$, the gradients of which define the directions of the paths at each latent code $\mathbf{z}\in\mathbb{R}^d$. More specifically, we transform $\mathbb{R}^d$ by $f^k\colon\mathbb{R}^d\to\mathbb{R}$ that is parameterized as a weighted sum of RBFs, and for any given $\mathbf{z}\in\mathbb{R}^d$ we move along the path belonging to the $k$-th family of paths by following the direction of $\nabla f^k(\mathbf{z})$. In order to obtain interpretable paths, we adopt the framework of~\cite{voynov2020unsupervised} and learn warping functions that give families of paths that lead to image transformations that are distinguishable to each other by a discriminator/reconstructor. The parameters of the warping function and of the reconstructor/discriminator network are optimized jointly. By contrast to~\cite{voynov2020unsupervised} and other methods in the literature, the warping functions may lead to non-linear paths, and the linear ones can be obtained for specific values of the parameters. An overview of the proposed method is given in Fig.~\ref{fig:overview}.
    
    \subsection{Vector space warping and traversal}\label{subsec:space_warping}
        Given a vector space $\mathbb{R}^d$, we define $f\colon\mathbb{R}^d\to\mathbb{R}$ as a weighted sum of parametric Gaussian RBFs given by
        \begin{equation}\label{eq:f}
            f(\mathbf{z}) = \sum_{i=1}^{N} \alpha_i \exp\left(-\gamma_i \lVert\mathbf{z}-\mathbf{s}_i\rVert^2\right), 
        \end{equation}
        where $\alpha_i\in\mathbb{R}$, $\gamma_i\in\mathbb{R}_+$, and $\mathbf{s}_i\in\mathbb{R}^d$, denote the weight, the scale, and the center of the $i$-th RBF, respectively. Geometrically, $f$ transforms each point $\mathbf{z}$ of the given vector space $\mathbb{R}^d$ into a $(d+1)$-dimensional point $(\mathbf{z},f(\mathbf{z}))$ that lies on a $d$-dimensional manifold. We define this transformation as a \textit{warping} of the vector space $\mathbb{R}^d$. Also, hereby, we will be referring to the centers of the RBFs as the \textit{support vectors}, driven by the geometric intuition that they ``support'' the induced warping of the space, and we will be using the term \textit{support set} to refer to the set of support vectors, $\mathcal{S}=\{\mathbf{s}_i\in\mathbb{R}^d,i=1,\ldots,N\}$. The corresponding weights and $\gamma$ parameters will be hereby referred to as the sets $\mathcal{A}=\{\alpha_i\in\mathbb{R},i=1,\ldots,N\}$ and $\mathcal{G}=\{\gamma_i\in\mathbb{R}_+,i=1,\ldots,N\}$, respectively. Then, different support sets will in general lead to different warpings of a given vector space. 
        
        The above warping operation is differentiable and its gradient is given analytically as follows
        \begin{equation}\label{eq:nabla_f}
            \nabla f(\mathbf{z}) = -2\sum_{i=1}^{N} \alpha_i\gamma_i \exp\left(-\gamma_i \lVert\mathbf{z}-\mathbf{s}_i\rVert^2\right)\left(\mathbf{z}-\mathbf{s}_i\right).
        \end{equation}
        Thus, given an arbitrary $\mathbf{z}$, $\nabla f(\mathbf{z})$ defines a (local) direction, which we use in order to define a curve in $\mathbb{R}^d$. More specifically, for any $\mathbf{z}\in\mathbb{R}^d$ and sufficiently small shift magnitude $\epsilon$, we define a continuous curve in $\mathbb{R}^d$ induced by the warping operation $f$ using (\ref{eq:nabla_f}) by shifting $\mathbf{z}$ by
        \begin{equation}\label{eq:deltaz}
            \delta\mathbf{z} = \epsilon\frac{\nabla f(\mathbf{z})}{\lVert\nabla f(\mathbf{z})\rVert}. 
        \end{equation}
         In Fig.~\ref{subfig:space_warping}, we illustrate this for a given vector space $\mathbb{R}^d$ and two warpings, $f^i$ and $f^j$, which lead to two different non-linear paths in $\mathbb{R}^d$ for any given $\mathbf{z}$ (dashed bold lines). In this figure, thin solid lines represent level sets of the warpings, while the vector fields represent their gradients.
         
        \begin{figure}[t]
            \centering
            \includegraphics[width=0.49\textwidth]{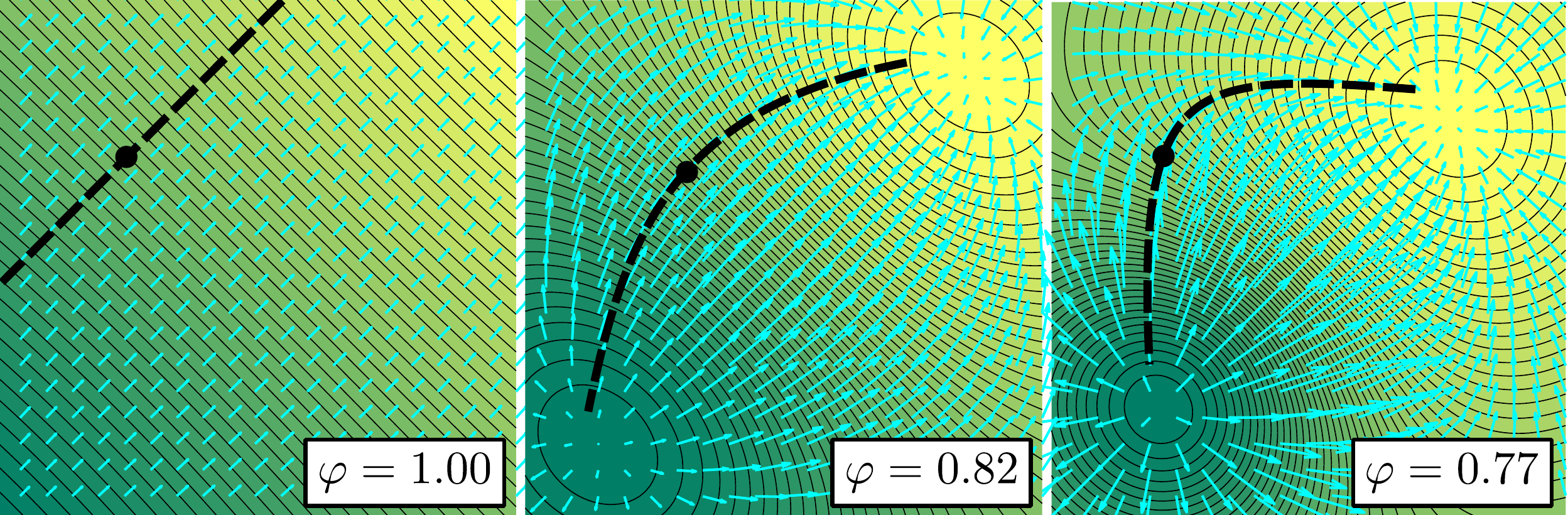}
            \caption{Illustration of gradient fields for warpings of bipolar RBFs, for small, medium, and large $\gamma$ parameter values.}
            \label{fig:bipolar}
        \end{figure}
    
    \subsection{Learning non-linear interpretable curves in GAN's latent space}
        Following the discussion above, given a pretrained GAN's latent space, which is typically modeled as a $d$-dimensional vector space $\mathcal{Z}\subseteq\mathbb{R}^d$, we may model a set of different warpings by a set of support sets $\{\mathcal{S}^k\}$, along with the corresponding weights $\{\mathcal{A}^k\}$ and $\gamma$ parameters $\{\mathcal{G}^k\}$, $k=1,\ldots,K$. We embed the support sets into the \textit{support tensor} $\mathbb{S}\in\mathbb{R}^{K\times N\times d}$, and the weights and $\gamma$ parameters into the matrices $\mathbb{A}\in\mathbb{R}^{K\times N}$ and $\mathbb{G}\in\mathbb{R}^{K\times N}$, respectively. Then, each support set, along with the corresponding weights and $\gamma$ parameters, leads to a specific warping of the latent space via the function $f^k$ defined by (\ref{eq:f}), whose gradient is given analytically by (\ref{eq:nabla_f}). Thus, for each $(\mathcal{S}^k,\mathcal{A}^k,\mathcal{G}^k)$, $k=1,\ldots,K$, we define a vector field on the latent space, which we use to traverse it using (\ref{eq:deltaz}). 
        
        Here, we define each warping to be given by a set of pairs of ``bipolar'' support vectors, i.e., pairs, that have opposite weights $\alpha$ and equal scale $\gamma$. In this formulation, $\gamma$ controls the degree of non-linearity of the path, where very small $\gamma$ lead to linear paths, similar to~\cite{voynov2020unsupervised}. This is illustrated in Fig.~\ref{fig:bipolar}, where the vector fields for two bipolar support vectors with different values of $\gamma$ are depicted.
        
        Finally, let us note that in contrast to the global linear directions discovered by~\cite{voynov2020unsupervised,ganspace_erik2020neurips}, in our case the directions along each warping are different for different latent codes. That is, as shown in (\ref{eq:deltaz}), the gradient and the shift vector depend on the latent code itself. This anisotropic behaviour of the proposed method reflects our intuition that interpretable paths do not necessarily have the same direction at every region of the latent space.
        
        \paragraph{Linear directions as a special case} In this section we will show that the method of~\cite{voynov2020unsupervised} can be derived as a special case of our method. We first note that the framework of~\cite{voynov2020unsupervised} that discovers linear directions encoded in the columns of a matrix $A$ can be derived in the special case that the warping functions are linear in $\mathbf{z}$, that is, $f(\mathbf{z})=A^\top\mathbf{z}$. In that case, the direction along the $k$-th direction is given by $\delta\mathbf{z}=\nabla f^k(\mathbf{z})=\mathbf{a}_k$, where $\mathbf{a}_k$ is the $k$-th column of $A$.
            
        It is straightforward to show that this solution can be obtained in our formulation, when each of the RBF-warping is given by pairs of bipolar RBFs, i.e, pairs of support vectors with opposite $\alpha$ and the same $\gamma$, when the value of $\gamma$ is sufficiently small. In what follows we give the proof for the simple case of a single bipolar pair, in the special case that $s_1 = -s_2 = s$. In that case, (\ref{eq:nabla_f}) can be written as $\nabla f^k(\mathbf{z}) = -2\alpha^k\gamma^k\exp\left(-\gamma^k\lVert\mathbf{z}-\mathbf{s}^k\rVert^2\right)\left(\mathbf{z}-\mathbf{s}^k\right)+2\alpha^k\gamma^k\exp\left(-\gamma^k\lVert\mathbf{z}+\mathbf{s}^k\rVert^2\right)\left(\mathbf{z}+\mathbf{s}^k\right)$, which, for sufficiently small $\gamma^k$, leads to $\nabla f^k(\mathbf{z}) = 4\alpha^k\gamma^k\mathbf{s}^k$. Then, the shift in the latent space, given by (\ref{eq:deltaz}), is written as
        \begin{equation}\label{eq:deltaz_linear}
            \delta\mathbf{z} 
            = \epsilon\frac{4\alpha^k\gamma^k\mathbf{s}^k}{\lVert4\alpha^k\gamma^k\mathbf{s}^k\rVert}
            = \epsilon\frac{\mathbf{s}^k}{\lVert\mathbf{s}^k\rVert}.
        \end{equation}
        In this case, the derivative of the $k$-th warping function at $\mathbf{z}$ is independent of $\mathbf{z}$ and equal to a constant vector.
            
        It is straightforward to show that linear directions can be obtained also in the more general case that each of the warping functions is given by several bipolar support vectors, each with a small $\gamma$. It is also the case that such parameters could be found by the optimization process, if they lead to discernible image transformations.
    
    \begin{figure}[t]
        \centering
        \includegraphics[width=0.48\textwidth]{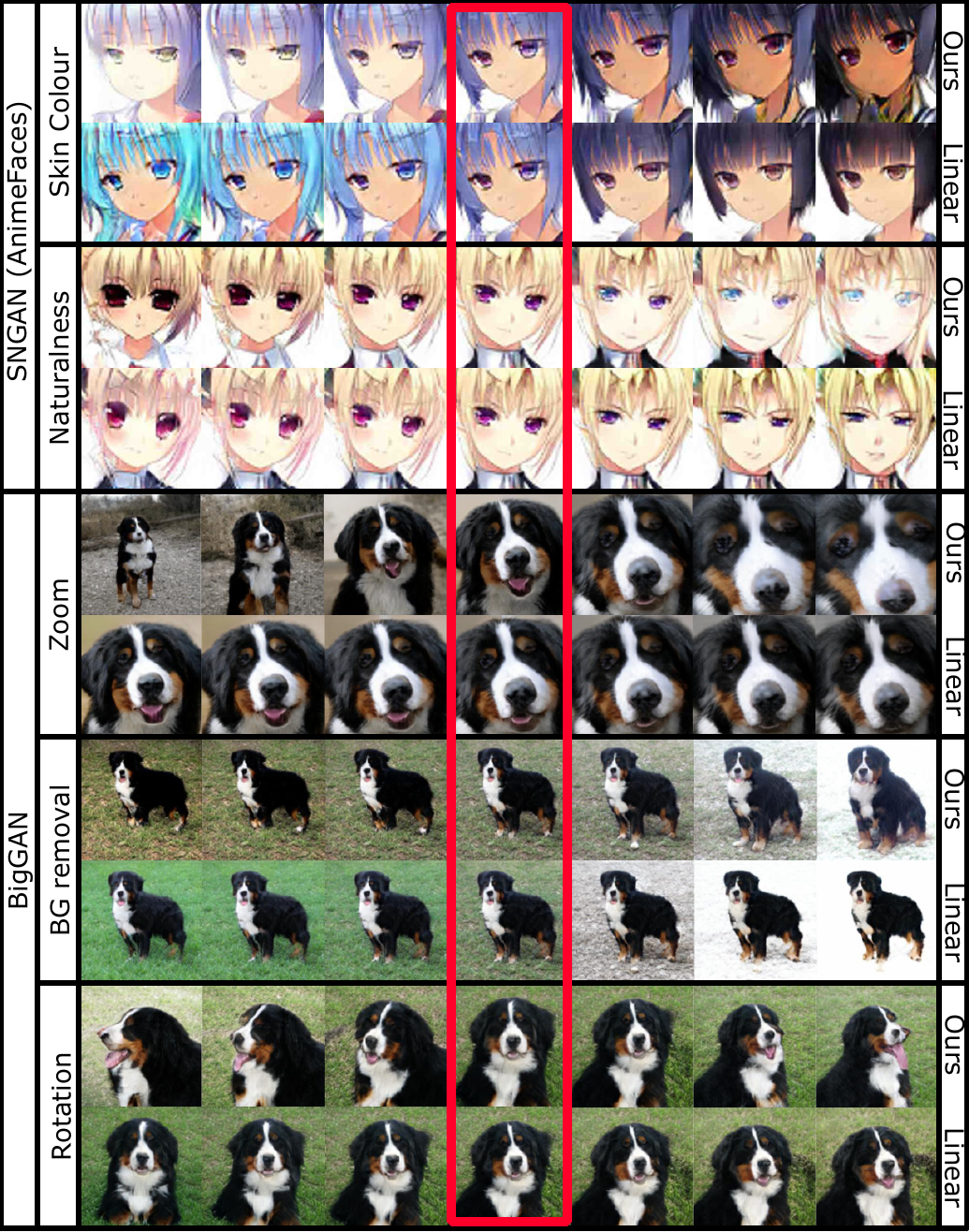}
        \caption{Interpretable paths discovered by our method (non-linear) compared to the corresponding linear ones discovered by~\cite{voynov2020unsupervised} for SN-GAN and BigGAN.}
        \label{fig:toy_gans}
    \end{figure}

    \subsection{Learning process}
        An overview of the learning process is presented in Fig.~\ref{fig:overview}. We use a pretrained generator \G and learn a) the parameters of a warping network $\mathcal{W}$ that generates paths in the latent space of \G, and b) the parameters of a reconstructor network \R that recognises the index $k$ of the warping that generated the changes between a pair of images. The trainable modules of our method are the following:
        
        \paragraph{Warping Network} The warping network $\mathcal{W}$ is parametrized by a set of triplets, $(\mathcal{S}^k,\mathcal{A}^k,\mathcal{G}^k)$, of the support set $\mathcal{S}^k$, and the corresponding weights $\mathcal{A}^k$ and $\gamma$ parameters $\mathcal{G}^k$, $k=1,\ldots,K$. Each such triplet gives rise to a warping of the latent space $\mathbb{R}^d$, and thus, to a non-linear path for any given latent code $\mathbf{z}\in\mathbb{R}^d$. $\mathcal{W}$ is implemented by standard layers and is differentiable.
        
        \paragraph{Reconstructor} A reconstructor \R is a model that we use in order to distinguish the image transformations that are induced by the different support sets (i.e., the different latent space warpings). As shown in Fig.~\ref{fig:overview}, the input to the reconstructor is a pair of images, $\mathfrak{G}(\mathbf{z})$ and $\mathfrak{G}\left(\mathbf{z}+\epsilon_k\frac{\nabla f^k(\mathbf{z})}{\lVert\nabla f^k(\mathbf{z})\rVert}\right)$. The reconstructor's goals are i) to predict which support set gave rise to the transformation at hand, i.e., recognise the index $k$ and ii) to reproduce the magnitude of the shift in the latent space; that is, predict $\epsilon_k$. In the experiments, we use the LeNet~\cite{lenet} backbone for SN-GAN (MNIST and AnimeFaces datasets) and ResNet-18~\cite{resnet_he2016cvpr} for BigGAN (ImageNet), ProgGAN (CelebA-HQ), and StyleGAN2 (FFHQ). We modify the input channels of the reconstructor so as it receives pairs of images (i.e., we concatenate the input image pair along channels dimension). Finally, we define two output ``heads'', one for predicting the index (classification), and the other for predicting the shift magnitude (regression). 
        
        \paragraph{Optimization objective} The optimization problem that we solve is as follows
        \begin{equation}\label{eq:min_loss}
            \min\limits_{\mathbb{S},\mathbb{A},\mathbb{G},\mathfrak{R}}
            \mathbb{E}_{\mathbf{z},k,\epsilon}
            \left[ \mathcal{L}_\text{cls}(k,\tilde{k}) + \lambda\mathcal{L}_\text{reg}(\epsilon,\tilde{\epsilon})
            \right],
        \end{equation}
        where $\mathcal{L}_\text{cls}$ denotes the classification loss term where we use the cross-entropy function, $\mathcal{L}_\text{reg}$ denotes the regression loss terms where we use the mean absolute error, and $\lambda$ is a weighting coefficient. We note that the objective function is differentiable with respect to the support vectors, weights $\alpha$ and $\gamma$ parameters, allowing us to learn not only the positions of the support vectors, but also their weights, and/or $\gamma$ parameters. To ensure the positivity of $\gamma$ we learn its logarithm. As discussed above, for each warping we learn a set of bipolar pairs of support vectors.
        
        During training, we generate pairs of images $\mathfrak{G}(\mathbf{z})$ and $\mathfrak{G}\left(\mathbf{z}+\epsilon_k\frac{\nabla f^k(\mathbf{z})}{\lVert\nabla f^k(\mathbf{z})\rVert}\right)$, where $\mathbf{z}\sim\mathcal{N}\left(\mathbf{0},I_d\right)$, $k$ is a warping function index uniformly sampled in $\{1,\dots, K\}$, and $\epsilon_k$ is a scalar sampled uniformly in $\mathcal{U}_{\mathcal{E}_{\min}}^{\mathcal{E}_{\max}}=\mathcal{U}[-\mathcal{E}_{\max},-\mathcal{E}_{\min}]\cup\mathcal{U}[\mathcal{E}_{\min},\mathcal{E}_{\max}]$. The pair of images is fed to the reconstructor where the loss is calculated and the gradients are back-propagated to the warping network and the reconstructor.

\section{Experiments}\label{sec:experiments}
    
    \begin{figure}[t]
        \centering
        \includegraphics[width=0.45\textwidth]{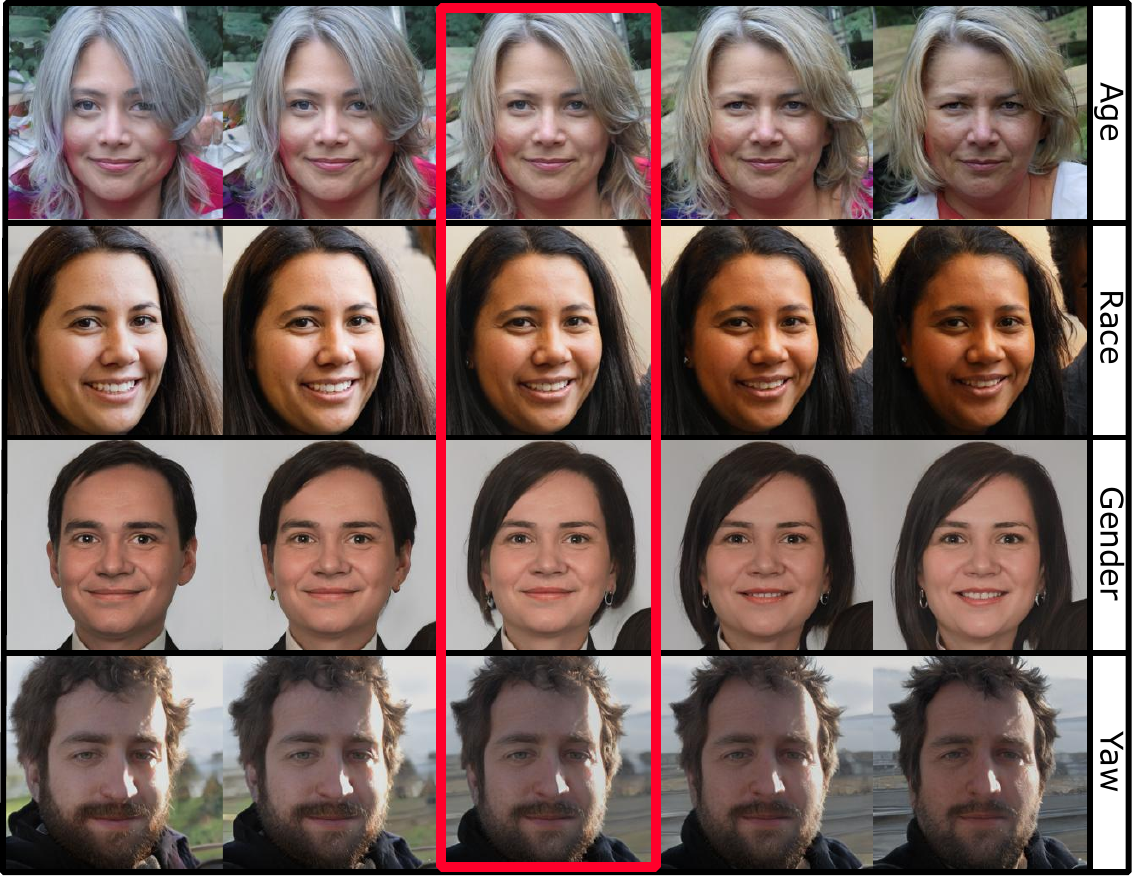}
        \caption{Non-linear interpretable paths automatically discovered by our method in StyleGAN2's~\cite{stylegan2_karras20cvpr} $\mathcal{W}$-space.}
        \label{fig:stylegan2_examples}
    \end{figure}
    
\begin{figure}[t]
        \centering
        \includegraphics[width=0.48\textwidth]{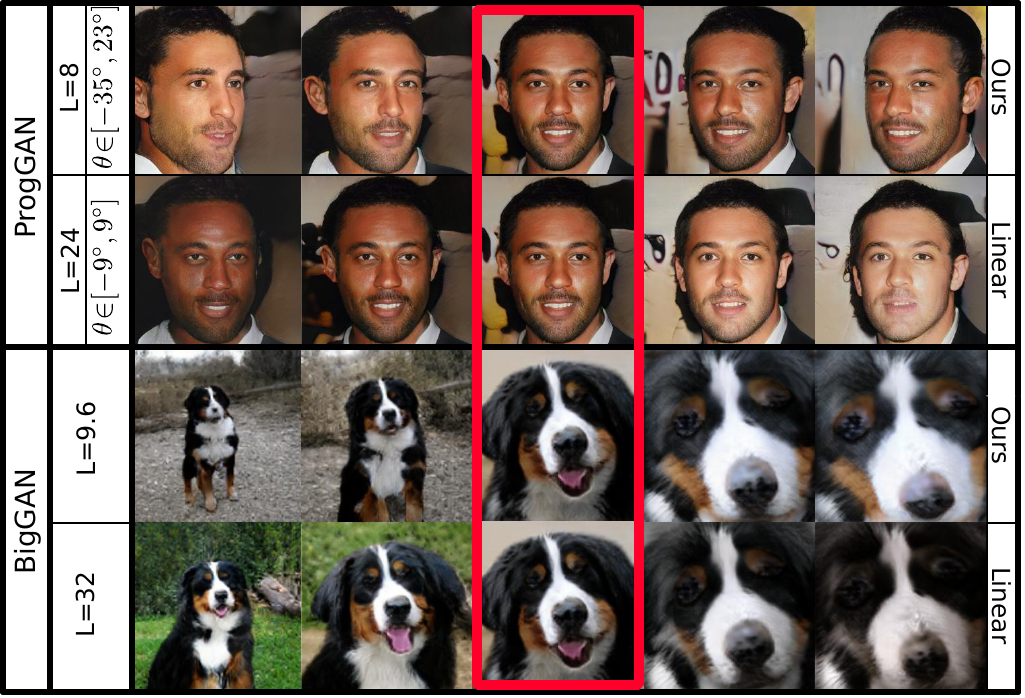}
        \caption{Effect of traversal path length $L$ in comparison with the linear case~\cite{voynov2020unsupervised}.} 
        \label{fig:length}
    \end{figure}

\begin{figure*}[t]
        \centering
        \includegraphics[width=0.88\textwidth]{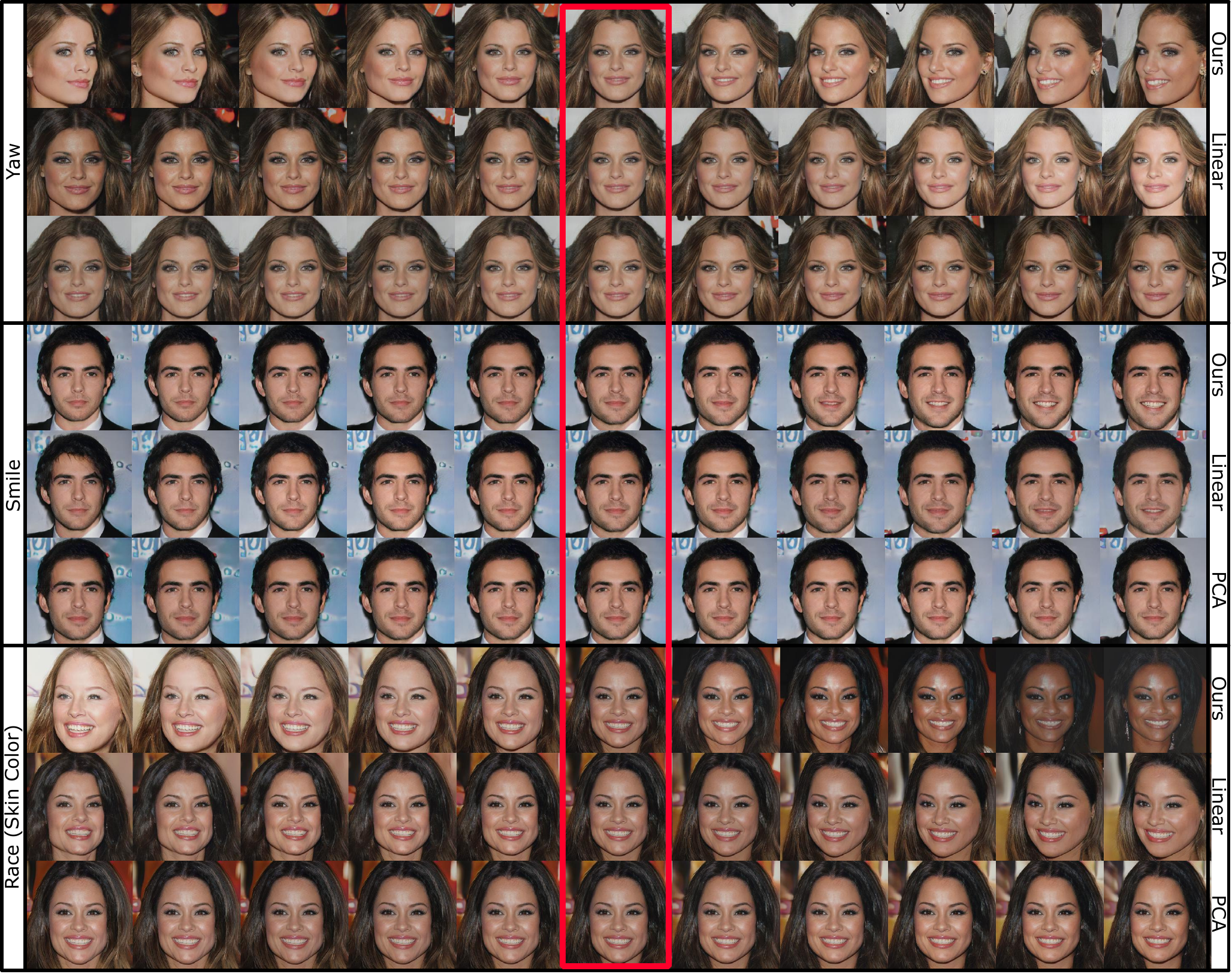}
        \caption{Automatically discovered non-linear (ours -- first row) and linear (Voynov and Babenko~\cite{voynov2020unsupervised} -- second row, GANSpace~\cite{ganspace_erik2020neurips} -- third row) interpretable paths in ProgGAN's~\cite{proggan_karras18iclr} latent space.}
        \label{fig:progan_results}
    \end{figure*}

    \paragraph{Overview of results} In this section we will present the experimental evaluation of the proposed method and provide qualitative and quantitative comparisons with state-of-the-art methods. We will first show that in comparison to~\cite{voynov2020unsupervised} our method finds paths in the latent space that produce changes in the image space that are easier to be distinguished by a discriminating network -- this is achieved consistently across several GANs that are pretrained on different datasets (Table~\ref{tab:accuracy}). We will then show that in comparison to the state-of-the-art, our method finds paths in the latent space, that produce more distinguishable, more disentangled and larger changes in the generated images. We will first show that qualitatively by presenting images generated along paths of equal length in the latent space for different methods (Fig.~\ref{fig:toy_gans},\ref{fig:progan_results},\ref{fig:stylegan2_examples}) and observe the generated variations they produce in the image space. We will subsequently show this quantitatively (Table~\ref{tab:proggan}), by estimating semantic attributes (e.g., rotations, smile, etc.) in the generated images, and report the correlations and ranges as we follow different paths in the latent space. Finally, we will show that our method finds paths on the latent space that correspond to steeper changes/paths in the attribute space, and therefore allows for better, controllable generation without arriving at latent space regions of low density and, thus, at quality degradation or distortions (Fig.~\ref{fig:length},\ref{fig:progan_results}).
    
    \begin{table}[t]
            \scriptsize
            \begin{tabular}{lccccc}
            \hline
            \multicolumn{1}{c}{\multirow{3}{*}{\begin{tabular}[c]{@{}c@{}}Method\end{tabular}}} & \multicolumn{5}{c}{GAN} \\ 
            \cline{2-6} 
            \multicolumn{1}{c}{} &  SNGAN  &  SNGAN  & \multirow{2}{*}{BigGAN} & \multirow{2}{*}{ProgGAN} & \multirow{2}{*}{StyleGAN2} \\
            \multicolumn{1}{c}{} & (MNIST) & (Anime) & & & \\ \hline\hline
            Random  & 46.0 & 85.0 & 76.0 & 60.0 & - \\
            Coord   & 48.0 & 89.0 & 66.0 & 82.0 & - \\
            Linear~\cite{voynov2020unsupervised} & 88.0 & 99.0 & 85.0 & 90.0 & - \\ \hline
            Ours & \textbf{98.4} & \textbf{99.8} & \textbf{92.6} & \textbf{99.3}  & \textbf{99.8} \\ \hline
            \end{tabular}
            \caption{Reconstructor accuracy (\%) of the proposed method compared to~\cite{voynov2020unsupervised} (linear directions), random latent direction and latent directions aligned with axes, for various GAN generators pretrained on the given datasets.}
            \label{tab:accuracy}
    \end{table}
    
    \begin{figure}[t]
        \centering
        \includegraphics[width=0.5\textwidth]{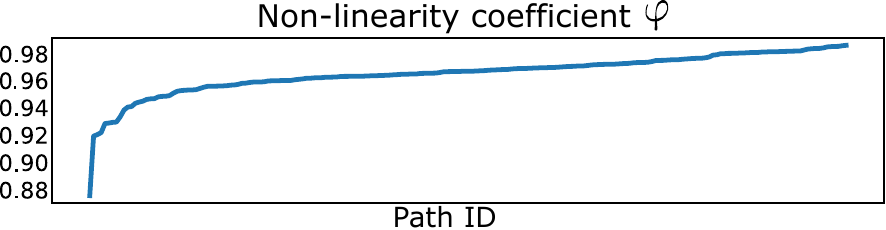}
        \caption{Illustration of the non-linearity coefficient $\varphi$ for the discovered paths obtained by our method for ProgGAN.}
        \label{fig:phi}
    \end{figure}
    
    \paragraph{Pretrained GAN generators and datasets} We evaluate the proposed method using the following pretrained GANs: a) Spectrally Normalized GAN (SN-GAN)~\cite{sngan_miyato18iclr} trained on MNIST~\cite{lecun1998mnist} and AnimeFaces~\cite{jin17neurips:animefaces}, b) BigGAN~\cite{biggan_brock19iclr} trained on ImageNet~\cite{deng2009cvpr:imagenet}, c) ProgGAN~\cite{proggan_karras18iclr} trained on CelebA-HQ~\cite{celeba_liu15iccv}, and d) StyleGAN2~\cite{stylegan2_karras20cvpr} trained on FFHQ~\cite{stylegan2_karras20cvpr}.
    
    \paragraph{Paths with more distinguishable changes in the image space} We first show that a reconstructor that discriminates images according to the warping in the latent space that generated them, i.e., estimates the index of the warping function, has better classification performance than in the corresponding linear case~\cite{voynov2020unsupervised}. This is an indication that the paths that are generated by our method can be discriminated more effectively and therefore are more likely to be more interpretable. The results are summarised in Table~\ref{tab:accuracy} and are consistent across several pretrained GANs.
    
    \begin{table}[ht]
        \footnotesize
        \centering
        \caption{Comparison of the proposed method (non-linear latent paths) to~\cite{voynov2020unsupervised} (linear latent directions) and GANSpace~\cite{ganspace_erik2020neurips} (linear PCA-based latent directions) in terms of $\mathcal{L}_1$-normalized correlation and range ($r$).}\label{tab:proggan}
        \begin{subtable}{0.48\textwidth}
            \centering
            \begin{tabular}{c||cccccc||c}
             & ID & Yaw & Pitch & Smile & Race & Hair & $r$ \\ \hline
            \multicolumn{1}{r||}{Yaw} & 0.52  & \textbf{0.32}  & 0.05  & 0.01  & 0.07  & 0.03   & \textbf{43.66$^\circ$} \\ \hline
            \multicolumn{1}{r||}{Pitch} & 0.41  & 0.04  & \textbf{0.38}  & 0.13  & 0.03  & 0.01 & \textbf{22.53$^\circ$} \\ \hline
            \multicolumn{1}{r||}{Smile} & 0.24  & 0.03  & 0.07  & \textbf{0.61}  & 0.03  & 0.03 & \textbf{0.37} \\ \hline
            \multicolumn{1}{r||}{Race} & 0.32  & 0.03  & 0.12  & 0.08  & \textbf{0.29}  & 0.17  & 0.06 \\ \hline
            \multicolumn{1}{r||}{Hair} & 0.23  & 0.02  & 0.11  & 0.13  & 0.02  & \textbf{0.49}  & \textbf{0.28} \\ \hline
            \end{tabular}
            \caption{Non-linear paths (Ours).}
            \label{tab:progan_ours}
        \end{subtable}
        \hfil
        \begin{subtable}{0.48\textwidth}
            \centering
            \begin{tabular}{c||cccccc||c}
             & ID & Yaw & Pitch & Smile & Race & Hair & $r$ \\ \hline
            \multicolumn{1}{r||}{Yaw} & 0.51  & \textbf{0.24}  & 0.21  & 0.01  & 0.02  & 0.01 & 18.93$^\circ$ \\ \hline
            \multicolumn{1}{r||}{Pitch} & 0.47  & 0.01  & \textbf{0.25}  & 0.04  & 0.00  & 0.22 & 8.27$^\circ$ \\ \hline
            \multicolumn{1}{r||}{Smile} & 0.24  & 0.01  & 0.04  & \textbf{0.57}  & 0.05  & 0.09 & 0.28 \\ \hline
            \multicolumn{1}{r||}{Race} & 0.52  & 0.05  & 0.02  & 0.10  & \textbf{0.31}  & 0.01 & \textbf{0.16} \\ \hline
            \multicolumn{1}{r||}{Hair} & 0.43  & 0.00  & 0.10  & 0.06  & 0.04  & \textbf{0.36} & 0.27 \\ \hline
            \end{tabular}
            \caption{Linear directions (Voynov and Babenko~\cite{voynov2020unsupervised}).}
            \label{tab:progan_linear}
        \end{subtable}
        \hfil
        \begin{subtable}{0.48\textwidth}
            \centering
            \begin{tabular}{c||cccccc||c}
             & ID & Yaw & Pitch & Smile & Race & Hair & $r$ \\ \hline
            \multicolumn{1}{r||}{Yaw} & 0.47  & \textbf{0.27}  & 0.04  & 0.13  & 0.03  & 0.06 & 17.65$^\circ$ \\ \hline
            \multicolumn{1}{r||}{Pitch} & 0.45  & 0.05  & \textbf{0.38}  & 0.09  & 0.02  & 0.01 & 7.48$^\circ$ \\ \hline
            \multicolumn{1}{r||}{Smile} & 0.21  & 0.00  & 0.07  & \textbf{0.55}  & 0.08  & 0.08 & 0.21 \\ \hline
            \multicolumn{1}{r||}{Race} & 0.35  & 0.11  & 0.02  & 0.12  & \textbf{0.27}  & 0.12 & 0.10 \\ \hline
            \multicolumn{1}{r||}{Hair} & 0.44  & 0.05  & 0.06  & 0.03  & 0.08  & \textbf{0.34} & 0.15 \\ \hline
            \end{tabular}
            \caption{Linear PCA directions (GANSpace~\cite{ganspace_erik2020neurips}).}
            \label{tab:progan_pca}
        \end{subtable}
    \end{table}
    
    \paragraph{Interpretable paths with steeper and more disentangled changes in the image space -- qualitative evaluation} We then show qualitatively that the proposed method finds interpretable paths in the latent space that are similar to the ones reported in~\cite{voynov2020unsupervised}, but exhibit larger variations in the captured generative factors. More specifically, for a given method that discovers a set of paths, that is, linear in the cases of~\cite{voynov2020unsupervised,ganspace_erik2020neurips} or non-linear in our case, in the latent space of a pretrained GAN, we generate an image sequence for each path, starting from a random latent code and ``walking'' towards the positive and the negative ways of the path for a certain amount of steps. This gives rise to an image sequence that shows how the learned path at hand affects the generation. For fair comparison, the step size and therefore the path length, is the same for all methods. 
    
    In Fig.~\ref{fig:toy_gans} we show the generated images along manually selected directions found by our method and the method of~\cite{voynov2020unsupervised} on SN-GAN (AnimeFaces). In the same figure, we show three interpretable paths discovered by our method, namely zoom, background removal, and rotation, in comparison with the corresponding ones reported in~\cite{voynov2020unsupervised} -- we note that these are the directions chosen in~\cite{voynov2020unsupervised} and that we generate the paths using the publicly available models provided by the authors. We can clearly see that in both cases, the paths found by our method produce larger changes in the image space and larger variations in the content.
    
    In Fig.~\ref{fig:progan_results} we show paths discovered on the latent space of ProgGAN~\cite{proggan_karras18iclr}, that is trained on CelebA-HQ~\cite{celeba_liu15iccv}. For this method we report the directions that are most correlated with three attributes, namely \textit{yaw}, \textit{smile}, and \textit{race}, with the correlations estimated with a method we will describe below. We compare with the corresponding linear directions obtained by~\cite{voynov2020unsupervised,ganspace_erik2020neurips} and we note that our method both leads to greater variation in the respective generative factors (e.g., larger rotation angles) for the same traversal lengths in the latent space, but also that we are able to produce more disentangled generations. This is apparent in Fig.~\ref{fig:progan_results}, where, for instance, changing \textit{smile} attribute using our method preserves other generative factors better than~\cite{voynov2020unsupervised,ganspace_erik2020neurips}.

    As noted, the length of the paths in the latent space is the same for all sequences and methods. To obtain a measure of the non-linearity of the generated paths, we calculate the ratio $\varphi$ between the length of a path and the distance between its endpoints, and report the averages for all the traversals on a given warping. Clearly, for linear paths, $\varphi=1$. The results are summarized in Fig.~\ref{fig:phi}, where we plot (sorted) the values of $\varphi$ for the discovered non-linear warpings for ProgGAN. An illustration is given in Fig.~\ref{fig:bipolar}.
    
    \paragraph{Non linear interpretable paths with steeper and more disentangled changes in the image space -- quantitative evaluation} In this section we will present our quantitative evaluation scheme, which we use for assessing the performance of our method and compare it to state-of-the-art~\cite{voynov2020unsupervised,ganspace_erik2020neurips}, for ProgGAN and StyleGAN2.
    
    As discussed before, for a given method that discovers a set of interpretable paths; that is, linear in the cases of~\cite{voynov2020unsupervised,ganspace_erik2020neurips} or non-linear in the case of the proposed method, in the latent space of a pretrained GAN generator, we generate an image sequence for each path, starting from a random latent code and ``walking'' towards the positive and the negative ways of the path for a certain amount of steps. For each image of such sequence, we apply a set of pretrained networks that predict the following: a) the location of the face (bounding box), using~\cite{zhang2017s3fd}, b) an identity score for each image of the sequence that expresses the similarity between the original image (central image of the sequence) and each of the rest, using ArcFace~\cite{deng2019arcface}, c) an age, race, and gender score using FairFace~\cite{karkkainen2019fairface}, d) a set of CelebA attributes classifiers (e.g., smile, wavy hair, etc.), and e) an estimation of the face pose (yaw, pitch, roll), using Hopenet~\cite{doosti2020hope}. In this way, for each warping we have a set of paths in the latent space and the corresponding paths in the attribute space.
    
    In order to obtain a measure on how well the paths generated by a warping function are correlated with a certain attribute, we estimate the average Pearson's correlation between the index of the step along the path and the corresponding values in the attribute vector. By doing so, for each warping, we obtain a vector, which we normalize. This allows for sorting the discovered paths with respect to the correlation with each attribute and select the paths that give the maximum absolute correlation for each attribute.
    
    The results are summarised in Table~\ref{tab:proggan}, where we report quantitative results for our method (Tab.~\ref{tab:progan_ours}), in comparison to~\cite{voynov2020unsupervised} (Tab.~\ref{tab:progan_linear}) and~\cite{ganspace_erik2020neurips} (Tab.~\ref{tab:progan_pca}), in terms of $\mathcal{L}_1$-normalized correlation averaged across 100 latent codes. We note that our method achieves better correlations for the respective attributes, while at the same time the correlations with the rest of the attributes are lower than those achieved by~\cite{voynov2020unsupervised,ganspace_erik2020neurips}, as is evident by the lower values in the off-diagonal elements of the matrix. This shows in a quantitative manner, what was evident in a qualitatively manner in Fig.~\ref{fig:progan_results}, that is, that the discovered paths in the latent space lead to more disentangled changes in the attribute space.
    
    Finally, in Fig.~\ref{fig:stylegan2_examples} we show the results of generation across some non-linear interpretable paths obtained automatically by our method for StyleGAN2, for the following attributes: \textit{age}, \textit{race} (skin color), \textit{gender} (``femaleness''), and \textit{yaw} (rotation). In this figure, we report the paths with the highest correlation with the respective attribute.

\section{Conclusion}\label{sec:conclusion}
    
    In this paper, we presented our method for discovering non-linear interpretable paths in the latent space of pretrained GANs in an unsupervised and model-agnostic manner. We do so by modeling non-linear latent paths using the gradient of RBF-based warping functions, which we optimized in order to be distinguishable to each other. This leads to paths that correspond to interpretable generation where only a small number of generative factors are affected for each path. Finally, we proposed a quantitative evaluation protocol for the case of face-generating GANs, which can be used to automatically associate the discovered paths with interpretable attributes such as smiling and rotation.

{\setlength{\parindent}{0cm}
\textbf{Acknowledgments:} This work was supported by the EU H2020 AI4Media No. 951911 project.
}

{\small
\bibliographystyle{ieee}
\bibliography{egbib}
}

\end{document}